\documentclass[conference]{IEEEtran}
\IEEEoverridecommandlockouts

\usepackage[a-1b]{pdfx}

\usepackage{hyperref}
\usepackage{cite}
\usepackage{amsmath,amssymb,amsfonts}
\usepackage{graphicx}
\usepackage{textcomp}
\usepackage{xcolor}
\usepackage{booktabs}
\usepackage{multirow}
\usepackage{xspace}
\usepackage{printlen}
\usepackage{bbm}
\usepackage{amsthm}

\usepackage[ruled, lined, linesnumbered, commentsnumbered, longend]{algorithm2e}

\usepackage{tikz,pgfplots}
\usetikzlibrary{tikzmark}
\usetikzlibrary{external}
\usetikzlibrary{pgfplots.groupplots}

\newtheorem*{remark}{Remark}

\newcommand{\HASTE}{TASER\xspace}

\newcommand{\STAB}[1]{\begin{tabular}{@{}c@{}}#1\end{tabular}}

\newcommand{\modify}[1]
{{\textcolor{black}{#1}}}

\def\BibTeX{{\rm B\kern-.05em{\sc i\kern-.025em b}\kern-.08em
    T\kern-.1667em\lower.7ex\hbox{E}\kern-.125emX}}
\begin{document}

\title{
TASER: \underline Temporal \underline Adaptive \underline Sampling for Fast and Accurat\underline{e} Dynamic Graph \underline Representation Learning
}

\makeatletter
\newcommand{\linebreakand}{%
  \end{@IEEEauthorhalign}
  \hfill\mbox{}\par
  \mbox{}\hfill\begin{@IEEEauthorhalign}
}
\makeatother

\author{
    \IEEEauthorblockN{Gangda Deng\IEEEauthorrefmark{1}\thanks{\IEEEauthorrefmark{1} Equal contribution}}
    \IEEEauthorblockA{
        \textit{University of Southern California} \\
        Los Angeles, USA \\
        gangdade@usc.edu
    }
    \and
    \IEEEauthorblockN{Hongkuan Zhou\IEEEauthorrefmark{1}}
    \IEEEauthorblockA{
        \textit{University of Southern California} \\
        Los Angeles, USA \\
        hongkuaz@usc.edu
    }
    \and
    \IEEEauthorblockN{Hanqing Zeng}
    \IEEEauthorblockA{
        \textit{Meta AI} \\
        Menlo Park, USA \\
        zengh@meta.com
    }
    \and
    \IEEEauthorblockN{Yinglong Xia}
    \IEEEauthorblockA{
        \textit{Meta AI} \\
        Menlo Park, USA \\
        yxia@meta.com
    }
    \linebreakand 
    \IEEEauthorblockN{Christopher Leung}
    \IEEEauthorblockA{
        \textit{Meta AI} \\
        Menlo Park, USA \\
        chrisleung@meta.com
    }
    \and
    \IEEEauthorblockN{Jianbo Li}
    \IEEEauthorblockA{
        \textit{Meta AI} \\
        Menlo Park, USA \\
        jianboli@meta.com
    }
    \and
    \IEEEauthorblockN{Rajgopal Kannan}
    \IEEEauthorblockA{
        \textit{US Army Research Lab} \\
        Los Angeles, USA \\
        rajgopal.kannan.civ@army.mil
    }
    \and
    \IEEEauthorblockN{Viktor Prasanna}
    \IEEEauthorblockA{
        \textit{University of Southern California} \\
        Los Angeles, USA \\
        prasanna@usc.edu
    }
}



\maketitle

\begin{abstract}
Recently, Temporal Graph Neural Networks (TGNNs) have demonstrated state-of-the-art performance in various high-impact applications, including fraud detection and content recommendation.
Despite the success of TGNNs, they are prone to the prevalent noise found in real-world dynamic graphs like time-deprecated links and skewed interaction distribution.
The noise causes two critical issues that significantly compromise the accuracy of TGNNs: (1) models are supervised by inferior interactions, and (2) noisy input induces high variance in the aggregated messages.
However, current TGNN denoising techniques do not consider the diverse and dynamic noise pattern of each node. 
In addition, they also suffer from the excessive mini-batch generation overheads caused by traversing more neighbors.
We believe the remedy for fast and accurate TGNNs lies in temporal adaptive sampling.
In this work, we propose \HASTE, the first adaptive sampling method for TGNNs optimized for accuracy, efficiency, and scalability. 
\HASTE adapts its mini-batch selection based on training dynamics and temporal neighbor selection based on the contextual, structural, and temporal properties of past interactions. 
To alleviate the bottleneck in mini-batch generation, \HASTE implements a pure GPU-based temporal neighbor finder and a dedicated GPU feature cache.
We evaluate the performance of \HASTE using two state-of-the-art backbone TGNNs. On five popular datasets, \HASTE outperforms the corresponding baselines by an average of 2.3\% in Mean Reciprocal Rank (MRR) while achieving an average of 5.1$\times$ speedup in training time.
\end{abstract}


\section{Introduction}
Dynamic graphs are natural abstractions of time-stamped interactions in many real-world systems. Interacting entities are represented as nodes, while interactions are represented as time-stamped edges.
Generating low-dimensional node representations on dynamic graphs (i.e., dynamic graph representation learning) is a fundamental problem for many practical systems, as it allows for monitoring and predicting the evolution of real-world data such as social networks, transportation networks, and financial networks.
Researchers have recently proposed various Temporal Graph Neural Networks (TGNNs)~\cite{evolvegcn, st-gcn, tgat, tgn, graphmixer, nat} to learn time-evolving patterns on dynamic graphs.
Unlike static graph representation learning approaches~\cite{gcn, sage, gat} that only accept time-invariant graphs as the input, TGNNs incorporate temporal information jointly with structural and contextual information into low-dimensional embeddings, which have shown superior performance in various real-world applications, including recommendation~\cite{zhang2022dynamic}, event prediction~\cite{zhou2021sedyt}, and fraud detection~\cite{9204584}.

Graph Neural Networks (GNNs), both static and temporal, recursively gather and aggregate information from neighboring nodes to generate node embeddings.
To reduce the high computation and memory footprint, neighbor sampling approaches~\cite{sage, saint, fastgcn, ladies} are widely used to alleviate the exponentially growing neighbor size with respect to the number of GNN layers.
However, most sampling methods approximate full neighborhood training using a static distribution, which is agnostic to the node/edge features, model architecture, and task performance. 
These sampling policies are vulnerable to noise since they cannot distinguish between relevant and irrelevant neighbors, leading to a large sampling variance.
To address these issues, researchers have designed adaptive sampling methods~\cite{as-gcn, bandit, pass, thanos}, where the sampling distribution is node-dependent and guided by the task performance.
On static graphs, these methods,  which come with theoretical guarantees for variance reduction and are adaptive to performance, can generate high-quality and robust node embeddings.

Dynamic graphs, much like their static counterparts, are not immune to the presence of noise, which adds false and irrelevant information to the graph signals. Specifically, dynamic graphs introduce two distinctive types of noise:
(1) \textbf{Deprecated links. }
Dynamic graphs accumulate an increasing number of interactions over time. Some old interactions could be irrelevant or even convey information that contradicts the current node status.
(2) \textbf{Skewed neighborhood distribution. }
The distribution of interactions among different nodes in a dynamic graph often exhibits significant disparities or sparsity.
Unlike static graphs, temporal graphs have many repeated edges between the same two nodes at different timestamps. 
A long-standing node may exhibit a skewed distribution of neighbors, while an emerging node may have very few neighbors.
For example, deprecated links can be observed in a social network graph when a person relocates to another country, rendering the previous connections gradually less informative or even incorrect.
Skewness becomes evident when an individual engages in daily conversations with their best friend while sending only a single message to a car dealer.
The noise in dynamic graphs causes two critical issues that significantly impair the accuracy of TGNNs. Firstly, when performing self-supervised training with the link prediction task, inferior interactions are used as positive links. Secondly, it is amplified in the iterative message-passing process, leading to high variance in the output embeddings.

To improve the performance of TGNNs on dynamic graphs with temporal and structural noise, researchers have proposed denoising techniques based on edge dropping~\cite{DDGCL,step} and implemented human-defined heuristics~\cite{tgat,tgn} to non-adaptive TGNN samplers. 
However, these approaches require extensive tuning and often achieve worse performance since they assume the whole graph follows the same noise pattern and ignore the differences in different nodes at different timestamps. 
For example, TGAT~\cite{tgat} employed the inverse timespan sampler to solve the deprecated links problem, which samples past neighbors with probabilities inversely proportional to their time deltas, but found that it performs worse than the original uniform sampler. 
Adaptive sampling, on the other hand, could learn customized sampling probabilities, encompassing any human-defined heuristics that may exist since the learnable sampler considers not only dynamic graph information but also training dynamics and task performance.
Therefore, we believe that adaptive sampling is integral to any approach addressing the noise problem in TGNNs, given its comprehensive consideration of all available information sources when estimating personalized neighborhood sampling probability distributions.

Despite the urgent need for adaptive sampling in TGNNs, it is notably challenging to construct an efficient and reliable solution. We identify the three main challenges as follows:
(1) To capture the dynamics in Temporal Graphs, the adaptive sampler should project not only structural and contextual information into sampling probabilities but also the time and frequency of the interactions.
(2) Existing adaptive sampling methods only support co-training with simple and static GNN aggregators and cannot be generalized to particularly complex temporal aggregators.
(3)Adaptive samplers require traversing a large and time-restricted neighborhood, resulting in enormous training time, especially when scaling to large-scale datasets.
Specifically, when the number of traversed neighbors increases, the mini-batch generation overheads (i.e., temporal neighbor finding and feature slicing with CPU-GPU data transfer) lead to an order-of-magnitude increase in the training time. Figure~\ref{fig:intro_tgat} shows the runtime breakdown of TGAT when the receptive field increases. On both datasets, the mini-batch generation time dominates the training time.
\modify{
Besides, adaptive sampling requires encoding node/edge features with learnable weights. Due to this compute-intensive nature, achieving fast adaptive sampling necessitates training on the GPU as well as specific GPU optimizations to alleviate the mini-batch generation bottleneck.
}

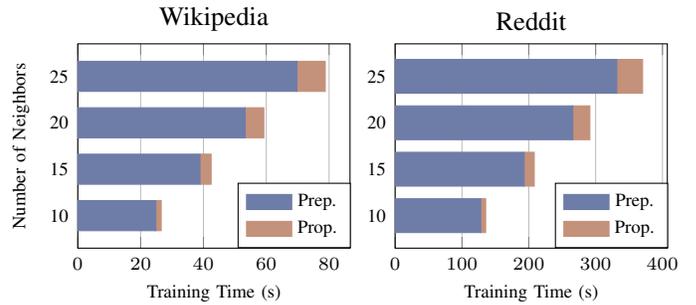
\begin{figure}[t]
\tikzsetnextfilename{teaser}
\begin{tikzpicture}




    \definecolor{c1}{RGB}{110, 124, 168}
    \definecolor{c2}{RGB}{196, 145, 122}


    \begin{groupplot}[
        group style={
            group size=2 by 1,
            horizontal sep=0.6cm,
        },
        tick label style={font=\scriptsize},
        label style={font=\scriptsize},
        width=5.2cm,
        height=4.3cm,
    ]

        \nextgroupplot[xbar stacked, bar width=4mm, title={Wikipedia},title style={yshift=-1ex}, xlabel={Training Time (s)},xlabel style={yshift=1ex},ylabel={Number of Neighbors},ylabel style={yshift=-3ex}, xmajorgrids,
            ymin=-0.7, ymax=3.7, ytick={0,1,2,3}, yticklabels={10,15,20,25}, ytick style={draw=none},
            xmin=0,
            legend style={font=\scriptsize, legend columns=1, at={(xticklabel cs:0.795)}, anchor=south, yshift=0.4cm}, area legend,
            ]
        
            \addplot[c1, fill=c1] table[x=wikiprep,y=idx,col sep=comma]{data/teaser.csv};
            \addplot[c2, fill=c2] table[x=wikiprop,y=idx,col sep=comma]{data/teaser.csv};
            \legend{Prep., Prop.}

        \nextgroupplot[xbar stacked, bar width=4.5mm, title={Reddit},title style={yshift=-1ex}, xlabel={Training Time (s)},xlabel style={yshift=1ex}, xmajorgrids,
            ymin=-0.7, ymax=3.7, ytick={0,1,2,3}, yticklabels={10,15,20,25}, ytick style={draw=none},
            xmin=0,
            legend style={font=\scriptsize, legend columns=1, at={(xticklabel cs:0.795)}, anchor=south, yshift=0.4cm}, area legend,
            ]
        
            \addplot[c1, fill=c1] table[x=redditprep,y=idx,col sep=comma]{data/teaser.csv};
            \addplot[c2, fill=c2] table[x=redditprop,y=idx,col sep=comma]{data/teaser.csv};
            \legend{Prep., Prop.}
            
    \end{groupplot}
\end{tikzpicture}%
\caption{Runtime (per epoch) breakdown for TGAT with different numbers of neighbors per layer. Prep. refers to the mini-batch generation time (neighbor finding, feature slicing, and CPU-GPU data transferring), while Prop. refers to the propagation time (forward and backward propagation).
} 
\label{fig:intro_tgat}
\end{figure}

To overcome the aforementioned challenges, we propose \HASTE, the first efficient and scalable adaptive sampling method for TGNNs.  
\HASTE provides a general solution for adaptive sampling in TGNNs and supports most TGNNs designed for Continuous Time Dynamic Graphs (CTDGs). 
To mitigate the mini-batch generation bottleneck, \HASTE implements a pure GPU-based temporal neighbor finder and a dedicated GPU feature cache.
Our main contributions are:
\begin{itemize}
    \item We propose a novel two-fold temporal adaptive sampling technique --- temporal adaptive mini-batch selection (Section~\ref{sec:method_batch_samp}) and temporal adaptive neighbor sampling (Section~\ref{sec:method_neigh_samp}). Temporal adaptive mini-batch selection selects high-quality training samples, while temporal adaptive neighbor sampling selects high-quality supporting neighbors.
    \item We implement the first GPU neighbor finder (Section~\ref{sec:method_scope}) for dynamic graphs, which is optimized for the massive Single Instruction Multiple Data (SIMD) GPU architecture.
    Compared with a state-of-the-art CPU neighbor finder, our GPU neighbor finder supports arbitrary training order while achieving an average speedup of $46\times$.
    \item We design a dynamic GPU cache (Section~\ref{sec:method_cache}) to speed up the feature-slicing process for large-scale datasets that cannot be fully stored on the GPU VRAM. Our cache replacing policy achieves near-optimal performance and requires minimal maintenance overhead. 
    \item In the experiments, we implement \HASTE on two state-of-the-art backbone TGNNs. On five popular datasets, \HASTE outperforms the corresponding baselines by an average of 2.3\% in MRR. Our GPU neighbor finder and 20\% GPU feature cache achieve an average of 5.1$\times$ speedup in the total training time.
\end{itemize}

\section{Background}

Dynamic graphs can be represented as a series of timestamped graph events. In this work, we consider the most common dynamic graphs with edges appearing as graph events. Without loss of generality, consider a dynamic graph $\mathcal{G}(\mathcal{V},\mathcal{E})$ with events $\{(u, v, \boldsymbol{x}_{uvt}, t)\}$, where each quadruplet represents an edge with edge feature $\boldsymbol{x}_{uvt}$ appearing from node $u$ to node $v$ at time $t$.
The goal of TGNNs is to generate information-rich dynamic embeddings for nodes at given timestamps.
The embeddings can be further used in different downstream tasks (e.g., clustering, node classification).
Note that TGNNs are usually trained self-supervised with the dynamic link prediction task (i.e., distinguish negative edges from positive edges)~\cite{tgn,tgat,graphmixer}. 
Let $\mathcal{E}_\mathcal{B}$ be a mini-batch of edges
sampled in the training set.
For each edge $(u, v, t) \in \mathcal{E}_\mathcal{B}$, a negative destination node $v'$ is randomly sampled from $\mathcal{V}$ to form a negative edge $(u,v',t)$.
Then, we sample supporting neighbors from the corresponding temporal neighborhood for these root nodes (i.e., $u$, $v$, and $v'$) and iteratively apply a series of temporal aggregators to compute their dynamic node embeddings.
Lastly, these embeddings are fed into an edge predictor where the binary cross entropy loss is used to perform an iteration of the Stochastic Gradient Descent (SGD) process.

\subsection{Neighbor Finder}
For node $v$ at time $t$, we consider its temporal neighborhood $\mathcal{N}(v, t) = \{(u, t_u) \mid  (v, u, t_u) \in \mathcal{E}, t_u < t\}$. 
To avoid the monotonically increased neighborhood size, TGNNs introduce a neighbor finder to select a subset of neighbors $\mathcal{N}_{s}(v, t) \subseteq \mathcal{N}(v, t)$ with a fixed size $b_{\mathcal{N}} = |\mathcal{N}_s(v, t)|$. 
Here, we introduce two neighbor finders that are widely used by existing TGNNs:
\begin{itemize}
    \item   
        Uniform Neighbor Finder ensures supporting nodes $\mathcal{N}_{s}(v, t)$ are sampled uniformly, deriving an unbiased approximation of the original neighborhood.
    \item
        Most Recent Neighbor Finder only samples the most recent neighbors, ensuring that the latest interactions between nodes are prioritized. 
\end{itemize}

\subsection{Temporal Aggregator} \label{sec:bg-tgnn}
For a target node $v$ at time $t$, its neighborhood $\mathcal{N}_s(v,t)$, and their input hidden features, a temporal aggregator performs the following two steps:
(1) encode time-aware embedding vector for each neighbor $(u, t_u) \in \mathcal{N}_s(v, t)$, and
(2) combine these vectors as the output vector.
To encode temporal information, existing methods map continuous timestamps to a $d_T$-dimensional vector space using $\boldsymbol{\Phi}: T \rightarrow \mathbb{R}^{d_T}$.
The $l$-th layer dynamic node embedding $\boldsymbol{h}_v^{(l)}$ for node $v$ at time $t$ can be computed as follows:

\begin{equation}
    \boldsymbol{m}_u^{(l)} = 
    \left\{ \boldsymbol{h}_u^{(l-1)}|| \boldsymbol{x}_{u v t} || \boldsymbol{\Phi} \left(\Delta t\right) \right\}
\end{equation}
\begin{equation}
    \boldsymbol{h}_v^{(l)} = \textup{COMB}\left(\boldsymbol{m}_v^{(l)}, \left\{\boldsymbol{m}_u^{(l)} \mid u \in \mathcal{N}_s(v, t) \right\}\right) 
\end{equation}
where $\Delta t=t-t_u$ and
\textup{COMB} is the combiner that aggregates all the related messages $\boldsymbol{m}_u^{(l)}$.
Note that if node $v$ interacts with node $u$ at different timestamps, we generate different messages sharing the same node embedding $\boldsymbol h_u^{(l-1)}$. 
Here, we introduce the two widely-used temporal aggregators:
\begin{itemize}
    \item
        TGAT ~\cite{tgat} is a self-attention-based aggregator that uses a learnable time encoding $\boldsymbol{\Phi}$:
        \begin{equation}
            \boldsymbol{\Phi} \left( \Delta t \right) = \text{cos}\left( \Delta t \boldsymbol{w} + \boldsymbol{b} \right),
        \end{equation}
        where $\boldsymbol{w} \in \mathbb{R}^{d_T}$ and $\boldsymbol{b} \in \mathbb{R}^{d_T}$ are learnable parameters.
        Let $\boldsymbol{M}_u^{(l)}$ be the $(l)$-th layer message matrix for temporal neighborhood $\mathcal{N}_s(v,t)$. The \textup{COMB} function of TGAT performs
        \begin{align}
            \boldsymbol{q}^{(l)}&=\boldsymbol{W}_q\left\{ \boldsymbol{h}_v^{(l-1)} \| \boldsymbol{\Phi}(0)\right\} + \boldsymbol{b}_s\\
            \boldsymbol{K}^{(l)}&=\boldsymbol{W}_k\boldsymbol{M}_u^{(l)}+\boldsymbol{b}_k\\
            \boldsymbol{V}^{(l)}&=\boldsymbol{W}_v\boldsymbol{M}_u^{(l)}+\boldsymbol{b}_v\\           \boldsymbol{h}_v^{(l)}&=\operatorname{Softmax}\left(\frac{\boldsymbol{q}^{(l)} \boldsymbol{K}^{(l)\mathrm{T}}}{\sqrt{\left|\mathcal{N}_s(v, t)\right|}}\right) \boldsymbol{V}^{(l)}. \label{eq:tgat_fw}
        \end{align}
    \item
        GraphMixer~\cite{graphmixer} provides a technically simple architecture with comparable performance to RNN-based and self-attention-based methods. It uses a fixed time-encoding 
        \begin{equation} \label{eq:time-encode}
            \boldsymbol{\Phi} \left( \Delta t \right) = \text{cos}\left( \Delta t \boldsymbol{\omega} \right),\quad\boldsymbol{\omega}=\left\{\alpha^{-(i-1) / \beta}\right\}_{i=1}^{d_T},
        \end{equation}
        followed by a 1-layer MLP-Mixer~\cite{mlp-mixer} aggregator to combine messages from neighboring nodes:
        \begin{equation}
            \boldsymbol{h}_v^{(l)} = 
            \operatorname{Mean}\left(\operatorname{MLP-Mixer}\left(
\boldsymbol{M}_u^{(l)}
            \right)\right).
        \end{equation}
\end{itemize}


\subsection{Related Works}

The recent success of practical GNN applications is attributed to their ability to quickly and accurately learn graph representations.
Techniques such as graph denoising and GPU accelerations enable GNNs to efficiently process noisy real-world data at large scales.


\noindent \textbf{Dynamic Graph Denoising.}
Existing dynamic graph denoising techniques are mainly based on dynamic graph sparsification.
TGAT~\cite{tgat} proposes a heuristics sampling policy based on the probability of inversed timespan.
TGN~\cite{tgn} further improves the timespan inversed sampling by sampling the most recent neighbors.
To avoid redundancy, TNS~\cite{tns} proposes to insert learnable spaces in the most recent neighbors.
STEP~\cite{step} proposes an unsupervised graph pruning method that drops the noisy interactions.
TGAC~\cite{tgac} devises dynamic graph augmentation techniques for contrastive TGNN learning, which measures the edge sample probability by computing the PageRank or Eigenvector of nodes in both ends.
However, all these methods do not consider the disparity of noise among different nodes at different times.

\noindent \textcolor{black}{ \textbf{Adaptive Mini-Batch Selection.}}
Adaptive mini-Batch selection, commonly referred to as adaptive importance sampling, constantly re-evaluates the relative importance of each training sample during training. The main idea behind these methods is to use gradient information to reduce variance in uniformly stochastic gradients in order to improve convergence. GRAD~\cite{zhu2016gradient} relies on both features and logits for solving least-squares problems. \cite{stich2017safe} proposed a general algorithm with efficient computation to speed up coordinate-descent and SGD. MVS~\cite{mvs} extends these methods to GNNs, considering both the variance introduced by mini-batch selection and neighbor sampling.
In contrast to reducing variance and speeding up optimization, TGNN training requires avoiding selecting noisy interactions as positive samples.

\noindent \textcolor{black}{ \textbf{Adaptive Neighbor Sampling.}}
As one of the graph denoising techniques, adaptive neighbor sampling methods learn a sample probability distribution for each neighboring node of a given target node.
AS-GCN~\cite{as-gcn} minimizes the GCN sampling variance by training a self-dependent function based on node features.
Bandit Sampling~\cite{bandit} formulates the variance reduction for adaptive sampling as an adversary bandit problem, and Thanos~\cite{thanos} further proposes a biased reward function to avoid instability.
In contrast to variance reduction, PASS~\cite{pass} directly optimizes task performance by approximating gradient propagation through the non-differentiable sampling operation of GCN.
To scale to large graphs, PASS adopts a two-step sampling approach, which first samples a fixed scope and then adaptively samples the neighbors within the scope.
However, these adaptive sampling methods can not capture the temporal information of dynamic graphs and are not compatible with temporal aggregators.


\noindent \textbf{Neighbor Finding.} 
Optimized GPU graph neighbor finders could achieve orders-of-magnitude higher throughput compared with CPU neighbor finders by leveraging the massive SIMD architecture and avoiding the data transfer overheads from CPU to GPU.
DGL~\cite{dgl} provides an easy-to-use GPU neighbor finder with unified virtual memory access support, demonstrating a speedup of 1.5$\times$ to 3.9$\times$ in total training time compared to the pipeline, which samples on the CPU and trains on the GPU.
Quiver~\cite{quiver} further proposes a workload-based scheduler that dynamically assigns tasks to the CPU and GPU to solve the imbalanced workload problem due to the unpredictable latency when working on sparse nodes.
Biased (weighted) neighbor finding based on inverse transformation sampling~\cite{c-saw}, rejection sampling~\cite{nextdoor}, and alias method~\cite{skywaker} are also well studied on GPUs.
However, they don't work on dynamic graphs and can not be used for temporal neighborhood sampling.
TGL~\cite{tgl} proposes the T-CSR data structure and a parallelized neighbor finder optimized for dynamic graphs on multi-core CPUs. Its key limitation is the reliance on pointer arrays for rapidly locating candidate temporal neighbors, which requires scheduling the training mini-batches chronologically.
Besides, Tea~\cite{tea} is a state-of-the-art general-purpose CPU random walk engine for biased neighbor finding on dynamic graphs. 
\modify{
However, it does not support high-dimensional feature transformation, which adaptive sampling requires.
}

\noindent \textbf{Graph Feature Caching.}
The neighbor explosion problem~\cite{sage} causes an enormous number of memory operations to fetch the node and edge features.
On large graphs whose entire node and edge feature matrices cannot be stored in GPU VRAM, the CPU-GPU feature slicing and loading process easily becomes the bottleneck during training.
GNS~\cite{gns} addresses this issue by periodically selecting a global set of nodes for all mini-batches and caching their features on GPU.
Data Tiering~\cite{data-tiering} uses reverse PageRank to predict the access probability of each node.
Quiver~\cite{quiver} further proposes a connectivity-aware node feature caching strategy that considers the probability of a node being sampled as a multi-hop neighbor.
However, these approaches are designed for the memory access pattern of static GNNs and do not consider temporal information.


\section{Approach}

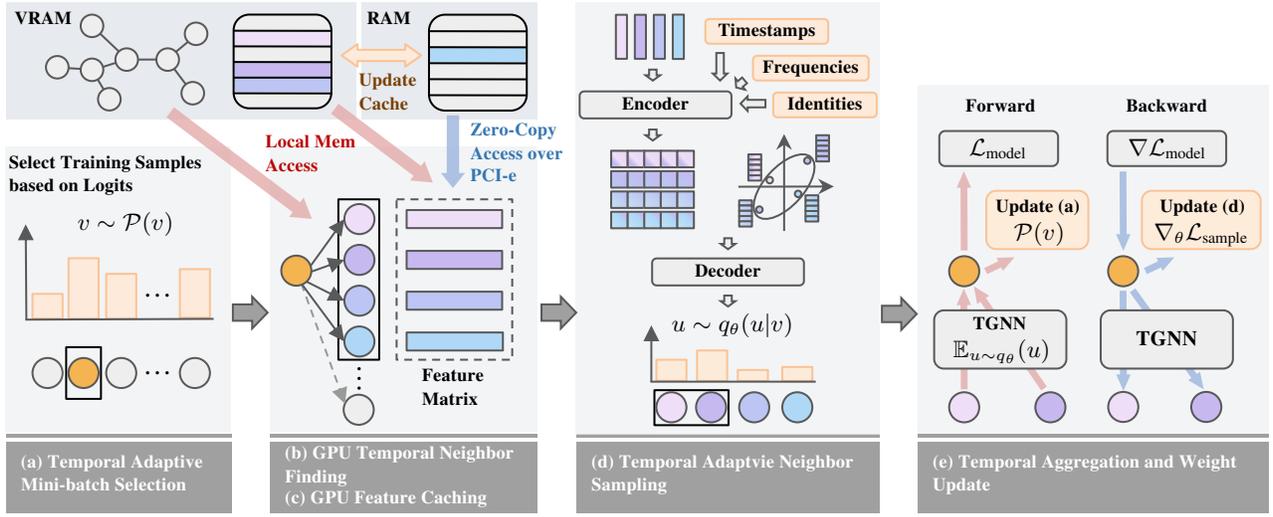
\begin{figure*}[t]
    \centering
    \tikzset{external/export next=false}
    \begin{tikzpicture}
        \node[inner sep=0pt] (plot) at (0,0){\includegraphics{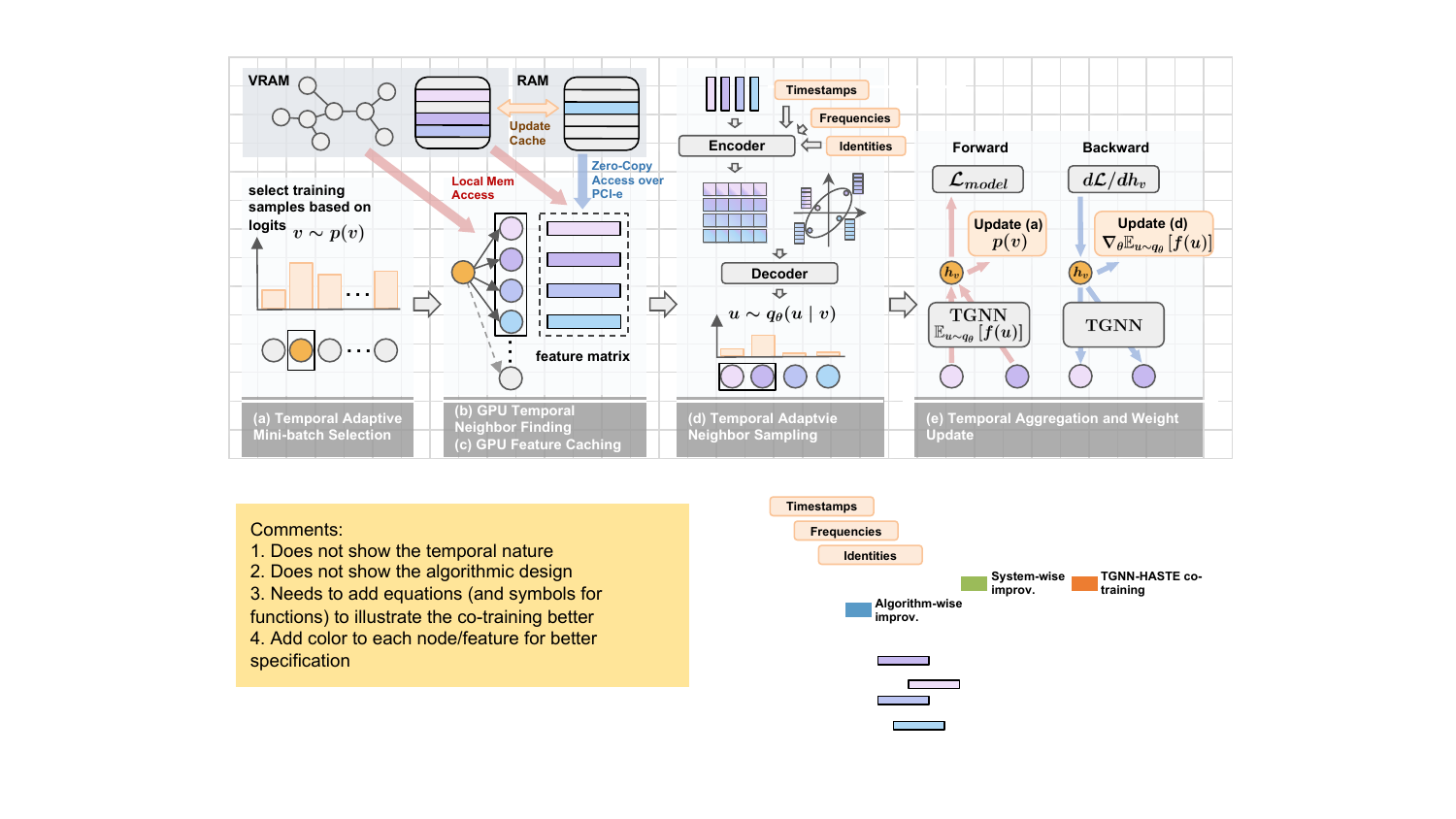}};
        \node[] at (-6.8,0.45) {\small$v\sim\mathcal P(v)$};
        \node[] at (1.27, -0.90) {\small$u \sim q_{\theta}(u|v)$};
        \node[] at (4.8, 1.45) {\small$\mathcal{L}_{\text{model}}$};
        \node[] at (7.05, 1.45) {\small$\nabla\mathcal{L}_{\text{model}}$};
        \node[] at (4.85, -1.25) {\small$\mathbb{E}_{u \sim q_{\theta}}(u)$};
        \node[] at (5.35, 0.32) {\small$\mathcal{P}(v)$};
        \node[] at (7.5, 0.3) {\small$\nabla_\theta\mathcal{L}_{\text{sample}}$};
    \end{tikzpicture}
    \caption{One training iteration of \HASTE on a one-layer TGNN. (a) Randomly select a set of mini-batch samples based on the pre-computed importance score $\mathcal{P}$ proportional to the logits (temporal adaptive mini-batch selection). (b) Sample a subset of neighbors from the temporal neighborhood using our GPU temporal neighbor finder. (c) Slice the features of sampled neighbors from the VRAM cache and RAM. (d) Apply temporal adaptive neighbor sampling (parameterized by $\theta$) to sub-sample the supporting neighbors for TGNN by encoding timestamps, frequencies, and identities along with features. (e) Perform forward and backward propagation. Update the importance score $\mathcal{P}$ for adaptive mini-batch selection and back-propagate through the model loss and sample loss to train the TGNN model and temporal adaptive sampler.} 
    \label{fig:method_overview}
\end{figure*}

In this section, we present \HASTE, a high-performance temporal adaptive sampling method for TGNNs. 
An overall illustration of one mini-batch training for \HASTE on a 1-layer TGNN is shown in \figurename~\ref{fig:method_overview}.
First, we maintain an importance score for each training sample, enabling the adaptive selection of a batch of high-quality samples in each step. 
Next, we adopt the bi-level neighbor sampling scheme used in PASS~\cite{pass} to improve the performance on large graphs. 
Initially, a GPU temporal neighbor finder samples a set of candidate neighbors from the temporal neighborhood $\mathcal{N}(v, t)$ using a static policy.
Then, we slice features from both the GPU cache and CPU memory, where the GPU cache is updated at the end of every epoch.
Following this, a parameterized temporal adaptive neighbor sampler is applied to sample a fixed-size set of informative supporting neighbors from the pre-sampled neighborhood. 
Finally, the TGNN model is trained on the representative node samples with their denoised supporting neighborhoods. 
We further update the sample importance score $\mathcal P(v)$ and the sampler's parameter $\theta$ during forward and backward propagation, respectively.

The rest of the section is arranged as follows.
We first propose our two-fold adaptive sampling technique regarding the mini-batch sample selection in Section~\ref{sec:method_batch_samp} and the supporting neighbor sampling in Section~\ref{sec:method_neigh_samp}. 
Then, we propose the pure-GPU neighbor finder in Section~\ref{sec:method_scope} and the dynamic GPU cache in Section~\ref{sec:method_cache}.

    
\subsection{Temporal Adaptive Mini-batch Selection} \label{sec:method_batch_samp}
In order to capture the pattern of node states changing over time, TGNNs are trained on interactions that cover the entire training set.
Unlike training GNNs on static graphs, where the models only recover the final states of different nodes, TGNNs need to recover different states for the same node during training.
However, learning to recover deprecated or cold-start states may significantly impair the accuracy of TGNNs.
To reduce the noise present in the training samples, we propose a temporal adaptive mini-batch selection method that utilizes the dynamic model predictions to sample high-quality training edges.

We first recall the original mini-batch SGD training process of TGNNs.
Given a training set $\mathcal{E}_{train}$, we chronologically sample a subset of edges $\mathcal{E}_\mathcal{B} \subseteq \mathcal{E}_{train}$ for each batch. 
For each training edge $e(v_1, v_2, t)\in\mathcal{E}_\mathcal{B}$, we set its label $y_e = 1$ and randomly sample a destination node $v_2'\in \mathcal{V}$ to form a negative edge $e'(v_1, v_2', t)$ with label $y_e'=0$.
During the forward propagation, we derive the logits $\hat{y}_e$ and $\hat{y}_{e'}$ for each positive and negative edge, respectively.
After that, the model loss is computed as follows:
\begin{equation}
    \mathcal{L}_{\textup{model}} = -\frac{1}{\left|\mathcal{E}_\mathcal{B}\right|} \sum_{e\in\mathcal{E}_\mathcal{B}}\phi(\hat{y}_e, y_e) + \phi(\hat{y}_{e'}, y_{e'}),
\end{equation}
where $\phi$ stands for the cross entropy loss function.

Unlike the original method that chronologically picks mini-batch samples, \HASTE adaptively selects mini-batch samples based on training dynamics.
First, we maintain a list of importance scores $\mathcal{P} \in \mathbb{R}^{|\mathcal{E}_{train}|}$ to evaluate the noisiness of each edge sample, which is initialized uniformly.
Then, we randomly sample a batch of training edges $\mathcal{E}_\mathcal{B}$ with the probability proportional to corresponding importance scores.
As shown in Figure~\ref{fig:method_overview}~(e), after the forward propagation, \HASTE update the importance score $\mathcal{P}(e)$ for every positive sample $e(v_1, v_2, t) \in \mathcal{E}_\mathcal{B}$ with
\begin{equation}
    \mathcal{P}(e) =\textup{sigmod}(\hat{y}_e) + \gamma,
\end{equation}
where $\gamma$ is a hyperparameter representing the magnitude of a uniform distribution mixed with the adaptive sample importance distribution.

Since dynamic graphs tend to be significantly noisier than static graphs, selecting positive samples with high confidence $\hat{y}_e$ can effectively improve accuracy.
Given the cross entropy loss function, the gradient of the loss with respect to logits is inversely proportional to the logits.
When the gradient update of a sample is large, it indicates that the sample is informative, but it is also more likely to be an outlier in the data distribution. 
To balance the noisiness and diversity in training samples, we can adjust the value of $\gamma$. A larger value of $\gamma$ makes the selector prone to sample noisier samples to amplify training. In practice, we found that $\gamma = 0.1$ works well on all the datasets.







\subsection{Temporal Adaptive Neighbor Sampling} \label{sec:method_neigh_samp}
Existing adaptive neighbor samplers only support attributed static graphs, which do not consider time restrictions and fail to distinguish recurring interactions in the temporal neighborhood.
They are specifically designed for a single type of aggregator and cannot achieve high accuracy when extended to other aggregators.
To address these issues, we propose a general encoder-decoder scheme that is suitable for various graph data and temporal aggregators.
Fig. \ref{fig:method_overview} (d) illustrates the forward propagation of our temporal adaptive neighbor sampler.
Formally, given a temporal neighborhood $\mathcal{N}_{s}(v_i, t_0)$, \HASTE  adaptively computes the sample policy $q_\theta\left(u_j, t_{k}|v_i, t_0\right)$ that estimates the probability of sampling neighbors $(u_j, t_k) \in \mathcal{N}_{s}(v_i, t_0)$ given node $v_i$ at time $t_0$. 

\noindent \textbf{Neighbor Encoder.}
Auxiliary information must be incorporated to discriminate the unique noise patterns in dynamic graphs, including outdated or redundant interactions.
To generate a time-aware sample policy, we use the fixed time-encoding function $TE_{\left( \Delta t \right)}$ proposed by GraphMixer~\cite{graphmixer} to encode temporal information for each neighbor, as shown in Eq. (\ref{eq:time-encode}).
$TE_{\left( \Delta t \right)}$ maps the relative timespan from the continuous time domain to a $d_{time}$-dimensional vector space. 
Besides being perceptual of time, the sampler also requires to distinguish the reappearance of neighboring nodes to differentiate redundant neighbors.
We propose the frequency encoding by leveraging the sinusoidal encoding~\cite{transformer}: 
\begin{equation}
    \begin{aligned}
F E_{\left(freq(u), 2 i\right)} & =\sin \left(freq(u) / 10000^{2 i / d_\text{freq}}\right) \\
F E_{\left(freq(u), 2 i-1\right)} & =\cos \left(freq(u) / 10000^{2 i / d_\text{freq}}\right)
\end{aligned}
\end{equation}
where $freq(u)$ denotes the frequency of a neighbor node $u$ appearing in the neighborhood $\mathcal{N}_{s}(v, t_0)$.
Since the frequency is indeed discrete and has limited values, we choose the positional encoding (i.e. sinusoidal encoding) instead of the time encoding to encode frequency.
However, when two nodes exhibit the same appearance frequency, the sampler remains unable to distinguish between them. To address this limitation, we propose the identity encoding $IE_{(u_j)}$. Given a sorted neighbor list $\{(u_1, t_1), (u_2, t_2), ..., (u_{b_\mathcal{N}}, t_{b_\mathcal{N}})\}$ of $\mathcal{N}_{s}(v, t_0)$, where $t_0 > t_1 > ... > t_{b_\mathcal{N}}$, we defined the identity encoding for each neighbor as:
\begin{equation}
    IE_{(u_j, i)} = \mathbbm{1}_{(u_j=u_i)},\quad i=1,2,\cdots,b_\mathcal N.
\end{equation}

In addition to these three encodings, we incorporate the contextual information of nodes and edges.
For each neighbor $(u, t) \in \mathcal{N}_{s}(v, t_0)$, we align the dimensions of node feature $\boldsymbol{x}_u$ and edge feature $\boldsymbol{x}_{vut}$ to $d_\text{feat}$:
\begin{equation}
    \boldsymbol{h}_{(u)} = \text{GeLU}(\boldsymbol{W_n}\boldsymbol{x}_u), ~ \boldsymbol{h}_{(v, u, t)} = \text{GeLU}( \boldsymbol{W_e}\boldsymbol{x}_{vut}).
\end{equation}
Finally, we concatenate all the encodings as well as features to derive the neighbor embedding as the input of the decoder:
\begin{equation}
    \boldsymbol{z}_{(u, t)} = \{\boldsymbol{h}_{(u)} || \boldsymbol{h}_{(v, u, t)} || TE_{(\Delta t)} || FE_{\left(freq(u)\right)} || IE_{(u)}\}.
\end{equation}
To ensure a balanced impact from various information sources, we set the dimensions to $d_\text{feat} = d_\text{time} = d_\text{freq}$ across all datasets. The dimension of neighbor embedding $\boldsymbol{z}_{(u, t)}$ is denoted as $d_{\text{enc}}$.


\noindent \textbf{Neighbor Decoder.}
After we encode the unique characteristics of dynamic graphs into the neighbor embedding $\boldsymbol{z}_{(u, t)}$, the subsequent processes can be modeled as a general adaptive neighbor sampling problem. 
For simplicity, we omit the timestamps and also do not differentiate the recurrence of interactions in a neighborhood.
The goal of the neighbor decoder is to generate a customized neighborhood importance distribution $q(u|v)$ for each neighborhood $\mathcal{N}_s(v)$.
Given that dynamic graphs often lack node features, rather than learning an exact pair-wise importance score for aggregation as GAT~\cite{gat}, we are more interested in estimating the relative importance of a node within the neighborhood $q(u|\{u', u' \in \mathcal{N}_s(v)\})$. 
Therefore, we use a 1-layer MLP-Mixer~\cite{mlp-mixer} to first transform the hidden embedding dimension and then transform the neighbor dimension for each neighborhood:
\begin{equation}
    \boldsymbol{Z}_{\mathcal{N}_s(v)} = \operatorname{MLP-Mixer}\left(\{\boldsymbol{z}_{u_1}, \boldsymbol{z}_{u_2}, ... , \boldsymbol{z}_{u_{b_\mathcal{N}}}\}\right),
\end{equation}
where $\boldsymbol{Z}_{\mathcal{N}_s(v)} \in \mathbb{R}^{b_\mathcal{N} \times d_\text{enc}}$.
In doing so, the neighbor embedding not only depends on the global transformation but also captures the neighborhood correlations.
To coordinate with different temporal aggregators, our neighbor decoder supports various predictors~\cite{gat, gatv2, transformer}, including
\begin{align}
            q_\text{linear}(u|v) &=            \sigma_u\left(\boldsymbol{w}_l\boldsymbol{Z}_{\mathcal{N}_s(v)} \right),\label{eq:decoder-linear}\\
            q_\text{gat}(u|v) &= 
\sigma_u\left(\text {LeakyReLU}\left(\boldsymbol{a}^\mathrm{T} \cdot\left[\boldsymbol{W}_g \boldsymbol{z}_u \| \boldsymbol{W}_g \boldsymbol{z}_v\right]\right) \right)\\
            q_{\text{gatv2}}(u|v) &= \sigma_u\left(\boldsymbol{a}^\mathrm{T}\text {LeakyReLU}\left( \boldsymbol{W}_{g2} \cdot \left[ \boldsymbol{z}_u \| \boldsymbol{z}_v\right]\right) \right)\label{eq:decoder-gat-v2}\\ 
            q_\text{trans}(u|v) &= \sigma_u\frac{\left(\boldsymbol{W}_t \boldsymbol{z}_v\right) \left( \boldsymbol{W}_t'\boldsymbol{Z}_{\mathcal{N}_s(v)}\right)^\mathrm{T}}{\sqrt{b_\mathcal{N}}}, \label{eq:decoder-trans}
\end{align}
where $\sigma$ is the softmax function. For the target node embedding $\boldsymbol z_v$, we concatenate the node feature (if it exists) with zero time encoding and one frequency encoding.
\begin{equation}
    \boldsymbol{z}_{v} = \{\boldsymbol{h}_{(v)} || TE_{(0)} || FE_{\left(1\right)}\}.
\end{equation}
Empirically, we observed that this target node embedding works well even without node features.


\begin{algorithm}[t]
    \caption{\HASTE Training: One Iteration} \label{alg:adapt-ns}
    \SetKwInOut{KwIn}{Require}
    \SetKwInOut{KwOut}{Ensure}

    \KwIn{a minibatch of labeled edges $\{e_k, y_k\}^b_{k=1}$, neighbor finding budget $m$, neighbor sampling budget $n$, $L$-layer TGNN model $f$, adaptive neighbor sampler $q(j|i)$}
    \KwOut{updated TGNN model and adaptive neighbor sampler}

     $V_{act}$ $\leftarrow$ all nodes in $\{e_k\}^b_{k=1}$\;
     
     $\mathcal{G}$ $\leftarrow$ empty supporting neighbor set\;
    
    \For{$l \leftarrow L$ \KwTo $1$}{
        \For{$(v_i, t) \in V_{act}$}{
        $\mathcal{N}_s$ $\leftarrow$ $\{u_j\}^{m}_{j=1}$ sampled from $\mathcal{N}(v_i, t)$\;
        
        $\mathcal{N}'_s$ $\leftarrow$ $\{u_j\}^{n}_{j=1}$ sampled from $\mathcal{N}_s$ with $q(j|i)$\;

        update $V_{act}$ and $\mathcal{G}_{[s]}$ with $\mathcal{N}'_s$\;
        }
    }

    $\mathcal{L}_\text{model}$ $\leftarrow$ loss($\{f(e_k, \mathcal{G}_{[k]}), y_k\}^{b}_{i=1}$)\;

    update $f$ by back-propagating $\mathcal{L}_\text{model}$\;

    $\mathcal{L}_\text{sample}$ $\leftarrow$ construct loss following Eq.(\ref{eq:tgat_bw}) or Eq.(\ref{eq:mxier_bw})\;

    update $q(j|i)$ by back-propagating $\mathcal{L}_\text{sample}$\;

    \KwRet{$f$ and $q(j|i)$}\;
\end{algorithm}

\noindent \textbf{Co-Training with Temporal Aggregators.}
Figure \ref{fig:method_overview} (e) demonstrates a temporal aggregator that combines the messages from sampled neighbors during the forward pass and subsequently back-propagates based on  $\mathcal L_\text{model}$. 
However, the parameter of the sampling policy $q_{\theta}(u|v)$ can not be directly updated through back-propagation since the sampling process is non-differentiable.
We need to construct an auxiliary loss $\mathcal{L}_\text{sample}$ to update $\theta$.
For an arbitrary temporal aggregator with neighbors sampled following $q_{\theta}(\cdot|v)$,  we can rewrite the forward propagation in the form of:
\begin{equation} \label{eq:expectation}
    \boldsymbol{h}_{v_i}^{(l)} = g^{(l)}\left(\left\{
    \mathbb{E}_{q_\theta(u_j|v_i)}\left[f\left(v_i, u_j\right)\right]
    , f \in \mathcal{H}^{(l)}
       \right\}\right),
\end{equation}
where $g^{(l)}$ and $\mathcal{H}^{(l)}$ are functions defined by the temporal aggregator at the $l$-th layer. 
This implies that every appearance of the expectation should be considered when calculating $\nabla_\theta \boldsymbol{h}_{v_i}^{(l)}$.
For simplicity, we concisely denote $q_\theta(u_j|v_i) $ as $ q_\theta(u_j)$ and $f(v_i, u_j) $ as $ f(u_j)$ below.
We can then approximate each $\nabla_\theta \mathbb{E}_{q_\theta(u_j)}[f(u_j)]$ using the log-derivative trick~\cite{williams1992simple} with $n$ Monte Carlo samples $\{u_j \sim q_\theta(u_j)\}_{j=1}^n$:
\begin{equation} \label{eq:log-derivative}
    \nabla_\theta\mathbb{E}_{q_\theta(u_j)}[f(u_j)]
\approx \frac{1}{n}\sum_{j=1}^n \nabla_\theta \log q_\theta(u_j) f(u_j).
\end{equation}
Next, we show how to calculate $\nabla_\theta \mathcal{L}_\text{model}$ when co-training with temporal aggregators.
For the TGAT aggregator shown in Eq. (\ref{eq:tgat_fw}), we denote $a_{i,j}$ as the unnormalized attention score between node $v_i$ and neighbor node $u_j$, and $\hat{a}_{i,j}=\operatorname{Softmax}(a_{i,j})$ is the normalized one.
The TGAT aggregator can be transformed in the form of Eq. (\ref{eq:expectation}) as: 
\begin{equation}
    \boldsymbol{h}_{v_i}^{(l)} = \mathbb{E}_{q_\theta(u_j)}\left[f_1(u_j)\right] ~ / ~ \mathbb{E}_{q_\theta(u_j)}\left[f_2(u_j)\right],
\end{equation}
where $f_1\left(u_j\right) = e^{a_{i,j}}[\boldsymbol{V}^{(l)}]_{u_j} $ and $f_2\left(u_j\right) = e^{a_{i,j}}$.
According to the chain rule and Eq. (\ref{eq:log-derivative}), we have:

\begin{equation}\label{eq:tgat_bw}
    \begin{aligned}
        &\nabla_\theta \mathcal{L}_\text{model} \\ &\approx \frac{d \mathcal{L}_\text{model}}{d \boldsymbol{h}_{v_i}^{(l)}} \cdot 
        \frac{1}{\lambda^{\alpha}}\sum_{j=1}^n \hat{a}_{i,j}\left(\left[\boldsymbol{V}^{(l)}\right]_{u_j} + \beta \boldsymbol{h}_{v_i}^{(l)}\right)\nabla_\theta \log q_\theta\left(u_j\right),
\end{aligned}
\end{equation}
where $\lambda = \mathbb{E}_{q_\theta\left(u_j\right)}\left[e^{a_{i,j}}\right]$. We additionally introduce two hyperparameters, $\alpha$ and $\beta$, to control the gradient variance and importance ratio between $v_i$ and $u_j$, respectively. In our experiments, we set $\alpha=2$ and $\beta=1$ for all datasets and find these values consistently perform well.

Similarly, for the GraphMixer aggregator, we have:
\begin{equation} \label{eq:mxier_bw}
    \nabla_\theta \mathcal{L}_\text{model} = \frac{d \mathcal{L}_\text{model}}{d \boldsymbol{h}_{v_i}^{(l)}} \cdot \frac{1}{n} \left\{    \sum\nolimits_{j=1}^n       w'_{jk} \mu_{jk} \nabla_\theta \log q_\theta(u_j)
    \right\}_{k=1}^{d_\text{enc}},
\end{equation}
where $\mu_{jk} = \boldsymbol{w}_k^{\mathrm{T}}\boldsymbol{h}_{u_j}^{(l-1)}$.

Based on Eq. (\ref{eq:tgat_bw}) and Eq.(\ref{eq:mxier_bw}), we can construct the sample loss $\mathcal{L}_\text{sample}$ by freezing the terms except for the log probability $\log q_\theta(u_j)$, and leveraging the autograd mechanism built in deep learning frameworks to update $\theta$.
Algorithm \ref{alg:adapt-ns} shows one iteration of \HASTE training on a $L$-layer TGNN. 

\begin{remark} (Adaptive sampling vs. Attention)
Contrary to the bottom-up approaches of graph attention, our temporal adaptive sampler computes in a top-down manner, which does not require any hidden features $h_u^{(l')}$ when generating sampling probabilities for nodes at layer $l$ ($l' < l$). Under a fixed-size scope, it reduces the computational complexity exponentially w.r.t. the number of layers.
\end{remark}


\begin{algorithm}[t]
    \caption{GPU Temporal Neighbor Finding} \label{alg:scope}
    \SetKwInOut{KwIn}{Input}
    \SetKwInOut{KwOut}{Output}
    \SetKwInOut{KwSetup}{Setup}
    
    \SetKwFor{ForParallel}{for}{do in parallel}{end}
    \SetKwFunction{BinarySearch}{BinarySearch}
    \SetKwFunction{CheckBitmap}{CheckBitmap}
    \SetKwFunction{SyncThreads}{SyncThreads}

    \KwIn{target nodes $\{(v_i, t'_{i})\}^b_{i=1}$, neighbor budget $m$, T-CSR graph $\mathcal{G}$}
    \KwOut{sampled neighbors $neigh$}
    
    \ForParallel{block $i \leftarrow 1$ \KwTo $b$}{
        \ForParallel{thread $j \leftarrow 1$ \KwTo $m$}{
            \If{$j = 1$}{            
                $\{(u_k, t_k)\}_{k=1}^{d_{v_i}} \leftarrow \mathcal{G}_{[v_i]}$\;
                $p \leftarrow$ \BinarySearch{$\{t_1, ..., t_{d_{v_i}}\}$,  $t'_{i}$}\;
                }
            
            \SyncThreads{}\;
            
            \uIf{most recent neighbor finding}{
                $neigh[i][j] \leftarrow$ $(u_{p-j}, t_{p-j})$\;
            }
            \uElseIf{uniform neighbor finding}{
                initialize bitmap $\mathcal{M}$\;
                \SyncThreads{}\;
                keep randomly selecting $r$ $ \in [1, p)$ until \CheckBitmap{$r$, $\mathcal{M}$} $= False$\;
                $neigh[i][j] \leftarrow$ $(u_r, t_r)$\;
            }
        }   
    }
\end{algorithm}

\subsection{GPU Temporal Neighbor Finding} \label{sec:method_scope}
Neighbor finding on dynamic Graphs is complex as the neighborhood $\mathcal{N}(v, t)$ for each node varies over time. 
To rapidly identify the candidate neighbor set, we store dynamic graphs in the T-CSR data structure~\cite{tgl}, which sorts the outgoing neighbors according to their timestamps.
As shown in Algorithm~\ref{alg:scope}, we employ a block-centric parallel sampling design to leverage the hierarchical GPU architecture.
Specifically, each target node is allocated a thread block, and each thread inside the block is assigned to sample a neighbor for the target node.
We first identify the pivot pointer in each neighborhood using binary search with a single thread and then use the shared memory bitmap~\cite{c-saw} for collision detection in uniform sampling without replacement.
After each thread selects a neighbor, an atomic compare-and-update operation is performed to detect whether this neighbor has been selected.
This block-centric design has three major benefits. 
Firstly, self-supervised TGNN training with \HASTE necessitates a large number of supporting neighbor candidates for thousands of mini-batch samples in each training iteration. The block-wise design can efficiently saturate GPU resources while avoiding intra-warp scheduling overhead.
Secondly, the threads in the same warp access the same neighbor information, which can be cached in the shared memory.
Thirdly, the complexity of the binary search is proportional to the neighbor size while the complexity of the bitmap is inversely proportional to the neighbor size, leading to a balanced workload across different blocks.
In experiment~\ref{sec:runtime}, we verify that our GPU neighbor finder achieves order-of-magnitude speedup compared with existing CPU neighbor finders~\cite{tgl}.


\begin{algorithm}[t]
    \caption{GPU Edge Feature Caching} \label{alg:cache}
    \SetKwInOut{KwIn}{Require}
    \SetKwInOut{KwOut}{Output}

    \KwIn{edge features $\{x_e, \forall e \in \mathcal{E}\}$, edge caching budget $k$, cache replacement threshold $\epsilon$}
    
    $\mathcal{Q} \leftarrow \{0\}_{i=1}^{|\mathcal{E}|}$\;
    Randomly cache $k$ edge features to VRAM\;
    
    \For{epoch $1$ \KwTo $T$}{
        \For{each edge read request $e$}{
                $\boldsymbol x_e$ serve from VRAM cache or RAM\;
                $\mathcal{Q}[e] \leftarrow  \mathcal{Q}[e] + 1$\;
            }
        \If{$|$cached edges $\cap$ $\mathcal{Q}_\text{topk}|< \epsilon$ }{        
                update cache with $\mathcal{Q}_{\text{top}k}$ edge features
        }
    }
\end{algorithm}

\begin{table*}[t!]
    \centering
    \caption{Accuracy of \HASTE and baselines in MRR (\%). All results are an average of 5 runs. (\textbf{First}, \underline{Second}.)}
    \setlength{\tabcolsep}{1.5pt}
    \renewcommand{\arraystretch}{1.2}
    \begin{tabular}{r||cc|cc|cc|cc|cc}
         & \multicolumn{2}{c|}{Wikipedia} & \multicolumn{2}{c|}{Reddit} & \multicolumn{2}{c|}{Flights} & \multicolumn{2}{c|}{MovieLens} & \multicolumn{2}{c}{GDELT}\\
         & TGAT & GraphMixer & TGAT & GraphMixer & TGAT & GraphMixer & TGAT & GraphMixer & TGAT & GraphMixer\\
        \toprule 
        Baseline & 68.76$\pm$0.40 & 74.05$\pm$0.21 & 81.05$\pm$0.04 & 75.11$\pm$0.07 & 80.50$\pm$0.08 & 77.86$\pm$0.08 & 63.11$\pm$0.04 & 69.01$\pm$0.06 & 79.01$\pm$0.12 & 76.26$\pm$0.47\\
        \midrule
        w./ Ada. Mini-Batch & 72.22$\pm$0.41 & \underline{75.34}$\pm$0.19 & \underline{82.57}$\pm$0.08 & \underline{76.25}$\pm$0.07 & \textbf{82.65}$\pm$0.06 & 78.89$\pm$0.09 & 63.97$\pm$0.08 & 69.11$\pm$0.04 & \underline{80.34}$\pm$0.04 & \underline{76.74}$\pm$1.09\\
        w./ Ada. Neighbor & \underline{73.96}$\pm$0.51 & 74.70$\pm$1.24 & 81.66$\pm$0.04 & 75.63$\pm$0.17 & 81.46$\pm$0.31 & \underline{78.94}$\pm$0.93 & \underline{65.51}$\pm$0.42 & \underline{69.26}$\pm$0.18 & 80.22$\pm$0.06 & 76.49$\pm$0.15\\
        \midrule
        \HASTE & \textbf{75.98}$\pm$0.35 & \textbf{76.48}$\pm$0.97 & \textbf{82.59}$\pm$0.16 & \textbf{76.85}$\pm$0.56 & \underline{82.64}$\pm$0.25 & \textbf{79.39}$\pm$0.64 & \textbf{65.79}$\pm$0.13 & \textbf{69.47}$\pm$0.10 & \textbf{81.04}$\pm$0.11 & \textbf{76.99}$\pm$0.72\\
        (Improvement) & (+7.22) & (+2.43) & (+1.54) & (+1.74) & (+2.14) & (+1.53) & (+2.68) & (+0.46) & (+2.03) & (+0.73)\\
    \end{tabular}
    \label{tab:mrr}
\end{table*}

\subsection{GPU Feature Caching} \label{sec:method_cache}
To address the issue of the dominant CPU-GPU feature slicing overhead in TGNN training, we propose a GPU cache for the features of nodes and edges with high-access frequency.
For dynamic graphs, since edge features are usually tremendously larger than node features, here we demonstrate the more commonly used case of edge feature caching.
Due to the temporal adaptive mini-batch selection and neighbor sampling, the access pattern of \HASTE changes during the training process, which requires a dynamic cache.
One naive approach is to maintain an $\mathcal{O}(|\mathcal{E}|\times |\mathcal{E}|)$ matrix to store the access frequency of every supporting neighbor for every training sample. However, this results in unacceptable storage overhead, and the cache update time may even exceed the training time.
Although increasing the cache line size can quadratically reduce the memory overhead, the cache hit rate also drops drastically due to the more coarse-grained policy. Empirically, we observe that increasing the cache line size from 1 to 512 leads to more than $20\%$ drop in cache hit rate.
On the other hand, since \HASTE uses the Adam optimizer, the dynamic edge access pattern will eventually stabilize.
Therefore, we leverage the historical edge access pattern to update the cache policy.
After each epoch, if the overlap between the cached edges and the $k$ most frequently accessed edges $\mathcal{Q}_\text{topk}$ of the previous epoch is less than a predefined threshold $\epsilon$, we swap the cached content with features of $\mathcal{Q}_\text{topk}$.
Note that this lightweight cache replacement policy only requires $\mathcal O(|\mathcal E|)$ computation, significantly less than the probability-based policy even with a large line size.



Algorithm~\ref{alg:cache} shows the GPU edge feature caching during training.
For each iteration of mini-batch training, \HASTE concurrently slices a batch of edge features layer by layer and updates the frequency of accessed edges in parallel.
For edge features that are not stored in the VRAM cache, we directly slice the feature through the unified virtual memory with zero-copy access over PCI-e.

\section{Experiments}

\begin{table}[b]
    \setlength{\tabcolsep}{0.9mm}
    \centering
    \caption{Dataset Statistic. $|d_v|$ and $|d_e|$ show the dimensions of node and edge features, respectively.} \label{tab:ds}
    \begin{tabular}{r|ccccccc}
        & $|\mathcal{V}|$ & $|\mathcal{E}|$ & $|d_v|$ & $|d_e|$ & train/val/test\\
        \toprule
        Wikipedia & 9,227 & 157,474 & - & 172 & 110k/23k/23k\\
        Reddit & 10,984 & 672,447 & - & 172 & 470k/101k/101k\\
        Flights & 13,169 & 1,927,145 & 100 & -& 600k/200k/200k\\
        MovieLens & 371,715 & 48,990,832 & - & 266& 600k/200k/200k\\
        GDELT & 16,682 & 191,290,882 & 413 & 130 & 600k/200k/200k\\
    \end{tabular}
\end{table}

\begin{table*}[t!]
    \centering
    \caption{Total Runtime Breakdown per Epoch (sec). NF, AS, FS, and PP denote neighbor finding, adaptive neighbor sampling, feature slicing, and propagation, respectively. The percentages (\%) represent the runtime ratios of a particular step relative to the total epoch. The arrows ($\uparrow$) refer to the same runtime as the one it points to.}
    \setlength{\tabcolsep}{3pt}
    \renewcommand{\arraystretch}{1}
    \begin{tabular}{r|r||cccc|c||cccc|c}
         & & \multicolumn{5}{c||}{TGAT} & \multicolumn{5}{c}{GraphMixer}\\
         & & NF (\%) & AS & FS (\%) & PP & Total (Impr.) & NF (\%) & AS & FS (\%) & PP & Total (Impr.)\\
        \toprule
        \multirow{5}{*}{\STAB{\rotatebox[origin=c]{90}{Wikipedia}}} & Baseline & 40.27 (70\%) & 2.\tikzmark{wiki2e}55 & 11.\tikzmark{wiki3e}26 (19\%) & 3.\tikzmark{wiki4e}73 & 57.81 (1.00$\times$) & 0.75 (23\%) & 0.\tikzmark{wiki6e}46 & 0.\tikzmark{wiki7e}61 (19\%) & 1.\tikzmark{wiki8e}45 & 3.28 (1.00$\times$)\\
         & +GPU NF & 0.\tikzmark{wiki1e}07 (0\%) & & \textcolor{white}{00.\tikzmark{wiki3s}00 }(64\%) & & 17.61 (3.28$\times$) & 0.\tikzmark{wiki5e}04 (2\%) & & \textcolor{white}{0.\tikzmark{wiki7s}00 }(24\%) & & 2.56 (1.27$\times$)\\
         & +10\% Cache & \textcolor{white}{0.00 }(1\%) & & 0.99 (13\%) & & 7.35 (7.86$\times$) & \textcolor{white}{0.00 }(2\%) & & 0.18 (8\%) & & 2.13 (1.53$\times$)\\
         & +20\% Cache & \textcolor{white}{0.00 }(1\%) & & 0.71 (10\%) & & 7.07 (8.17$\times$) & \textcolor{white}{0.00 }(2\%) & & 0.16 (8\%) & & 2.11 (1.55$\times$)\\
         & +30\% Cache & \textcolor{white}{0.\tikzmark{wiki1s}00 }(1\%) & \textcolor{white}{0.\tikzmark{wiki2s}00} & 0.54 (8\%) & \textcolor{white}{0.\tikzmark{wiki4s}00} & 6.90 (8.38$\times$) & \textcolor{white}{0.\tikzmark{wiki5s}00 }(2\%) & \textcolor{white}{0.\tikzmark{wiki6s}00} & 0.13 (6\%) & \textcolor{white}{0.\tikzmark{wiki8s}00} & 2.08 (1.57$\times$)\\
        \midrule
        \multirow{5}{*}{\STAB{\rotatebox[origin=c]{90}{Reddit}}} 
         & Baseline & 218.56 (77\%) & 10.\tikzmark{reddit2e}18 & 41.\tikzmark{reddit3e}56 (15\%) & 12.\tikzmark{reddit4e}62 & 282.93 (1.00$\times$) & 3.23 (23\%) & 1.\tikzmark{reddit6e}98 & 2.\tikzmark{reddit7e}36 (17\%) & 6.\tikzmark{reddit8e}23 & 13.79 (1.00$\times$)\\
         & +GPU NF & 0.\tikzmark{reddit1e}37 (1\%) & & \textcolor{white}{00.\tikzmark{reddit3s}00 }(64\%) & & 64.73 (4.37$\times$) & 0.\tikzmark{reddit5e}19 (2\%) & & \textcolor{white}{0.\tikzmark{reddit7s}00 }(22\%) & & 10.75 (1.28$\times$)\\
         & +10\% Cache & \textcolor{white}{0.00 }(1\%) & & 4.41 (16\%) & & 27.58 (10.25$\times$) & \textcolor{white}{0.00 }(2\%) & & 0.81 (9\%) & & 9.19 (1.50$\times$)\\
         & +20\% Cache & \textcolor{white}{0.00 }(1\%) & & 2.95 (11\%) & & 26.12 (10.82$\times$) & \textcolor{white}{0.00 }(2\%) & & 0.71 (8\%) & & 9.09 (1.51$\times$)\\
         & +30\% Cache & \textcolor{white}{0.\tikzmark{reddit1s}00 }(2\%) & \textcolor{white}{00.\tikzmark{reddit2s}00} & 2.36 (9\%) & \textcolor{white}{00.\tikzmark{reddit4s}00} & 25.53 (11.08$\times$) & \textcolor{white}{0.\tikzmark{reddit5s}00 }(2\%) & \textcolor{white}{0.\tikzmark{reddit6s}00} & 0.60 (7\%) & \textcolor{white}{0.\tikzmark{reddit8s}00} & 8.98 (1.53$\times$)\\
        \midrule
        \multirow{5}{*}{\STAB{\rotatebox[origin=c]{90}{MovieLens}}} & Baseline & 276.28 (69\%) & 17.\tikzmark{movielens2e}62 & 79.\tikzmark{movielens3e}15 (20\%) & 27.\tikzmark{movielens4e}60 & 400.66 (1.00$\times$) & 5.57 (13\%) & 4.\tikzmark{movielens6e}80 & 19.\tikzmark{movielens7e}61 (52\%) & 12.\tikzmark{movielens8e}71 & 42.68 (1.00$\times$)\\
         & +GPU NF & 0.\tikzmark{movielens1e}54 (0\%) & & \textcolor{white}{00.\tikzmark{movielens3s}00 }(63\%) & & 124.92 (3.20$\times$) & 0.\tikzmark{movielens5e}32 (1\%) & & \textcolor{white}{00.\tikzmark{movielens7s}00 }(24\%) & & 37.45 (1.14$\times$)\\
         & +10\% Cache & \textcolor{white}{0.00 }(1\%) & & 12.05 (21\%) & & 57.81 (6.93$\times$) & \textcolor{white}{0.00 }(2\%) & & 0.46 (2.5\%) & & 18.30 (2.33$\times$)\\
         & +20\% Cache & \textcolor{white}{0.00 }(1\%) & & 9.67 (17\%) & & 55.43 (7.22$\times$) & \textcolor{white}{0.00 }(2\%) & & 0.45 (2.5\%) & & 18.29 (2.33$\times$)\\
         & +30\% Cache & \textcolor{white}{0.\tikzmark{movielens1s}00 }(1\%) & \textcolor{white}{00.\tikzmark{movielens2s}00} & 7.99 (15\%) & \textcolor{white}{00.\tikzmark{movielens4s}00} & 53.75 (7.45$\times$) & \textcolor{white}{0.\tikzmark{movielens5s}00 }(2\%) & \textcolor{white}{0.\tikzmark{movielens6s}00} & 0.46 (2.5\%) & \textcolor{white}{00.\tikzmark{movielens8s}00} & 18.30 (2.33$\times$)\\
        \midrule
        \multirow{5}{*}{\STAB{\rotatebox[origin=c]{90}{GDELT}}} & Baseline & 322.40 (83\%) & 17.\tikzmark{gdelt2e}08 & 17.\tikzmark{gdelt3e}84 (5\%) & 29.\tikzmark{gdelt4e}52 & 386.84 (1.00$\times$) & 6.5 (12\%) & 6.\tikzmark{gdelt6e}19 & 15.\tikzmark{gdelt7e}37 (29\%) & 25.\tikzmark{gdelt8e}33 & 53.33 (1.00$\times$)\\
         & +GPU NF & 0.\tikzmark{gdelt1e}56 (1\%) & & \textcolor{white}{000.\tikzmark{gdelt3s}00 }(27\%) & & 65.00 (5.95$\times$) & 0.\tikzmark{gdelt5e}36 (1\%) & & \textcolor{white}{00.\tikzmark{gdelt7s}00 }(32\%) & & 47.24 (1.12$\times$)\\
         & +10\% Cache & \textcolor{white}{0.00 }(1\%) & & 3.08 (6\%) & & 50.25 (7.69$\times$) & \textcolor{white}{0.00 }(1\%) & & 0.52 (2\%) & & 32.40 (1.64$\times$)\\
         & +20\% Cache & \textcolor{white}{0.00 }(1\%) & & 2.17 (4\%) & & 49.34 (7.83$\times$) & \textcolor{white}{0.00 }(1\%) & & 0.54 (2\%) & & 32.42 (1.64$\times$)\\
         & +30\% Cache & \textcolor{white}{0.\tikzmark{gdelt1s}00 }(1\%) & \textcolor{white}{00.\tikzmark{gdelt2s}00} & 2.39 (5\%) & \textcolor{white}{00.\tikzmark{gdelt4s}00} & 49.56 (7.80$\times$) & \textcolor{white}{0.\tikzmark{gdelt5s}00 }(1\%) & \textcolor{white}{0.\tikzmark{gdelt6s}00} & 0.54 (2\%) & \textcolor{white}{00.\tikzmark{gdelt8s}00} & 32.41 (1.64$\times$)\\
    \end{tabular}
    \tikzset{external/export next=false}
    \begin{tikzpicture}[overlay, remember picture, shorten >=0.1cm]
        \draw [->] ({pic cs:wiki1s}) to ({pic cs:wiki1e});
        \draw [->] ({pic cs:wiki2s}) to ({pic cs:wiki2e});
        \draw [->] ({pic cs:wiki3s}) to ({pic cs:wiki3e});
        \draw [->] ({pic cs:wiki4s}) to ({pic cs:wiki4e});
        \draw [->] ({pic cs:wiki5s}) to ({pic cs:wiki5e});
        \draw [->] ({pic cs:wiki6s}) to ({pic cs:wiki6e});
        \draw [->] ({pic cs:wiki7s}) to ({pic cs:wiki7e});
        \draw [->] ({pic cs:wiki8s}) to ({pic cs:wiki8e});

        \draw [->] ({pic cs:reddit1s}) to ({pic cs:reddit1e});
        \draw [->] ({pic cs:reddit2s}) to ({pic cs:reddit2e});
        \draw [->] ({pic cs:reddit3s}) to ({pic cs:reddit3e});
        \draw [->] ({pic cs:reddit4s}) to ({pic cs:reddit4e});
        \draw [->] ({pic cs:reddit5s}) to ({pic cs:reddit5e});
        \draw [->] ({pic cs:reddit6s}) to ({pic cs:reddit6e});
        \draw [->] ({pic cs:reddit7s}) to ({pic cs:reddit7e});
        \draw [->] ({pic cs:reddit8s}) to ({pic cs:reddit8e});

        \draw [->] ({pic cs:movielens1s}) to ({pic cs:movielens1e});
        \draw [->] ({pic cs:movielens2s}) to ({pic cs:movielens2e});
        \draw [->] ({pic cs:movielens3s}) to ({pic cs:movielens3e});
        \draw [->] ({pic cs:movielens4s}) to ({pic cs:movielens4e});
        \draw [->] ({pic cs:movielens5s}) to ({pic cs:movielens5e});
        \draw [->] ({pic cs:movielens6s}) to ({pic cs:movielens6e});
        \draw [->] ({pic cs:movielens7s}) to ({pic cs:movielens7e});
        \draw [->] ({pic cs:movielens8s}) to ({pic cs:movielens8e});

        \draw [->] ({pic cs:gdelt1s}) to ({pic cs:gdelt1e});
        \draw [->] ({pic cs:gdelt2s}) to ({pic cs:gdelt2e});
        \draw [->] ({pic cs:gdelt3s}) to ({pic cs:gdelt3e});
        \draw [->] ({pic cs:gdelt4s}) to ({pic cs:gdelt4e});
        \draw [->] ({pic cs:gdelt5s}) to ({pic cs:gdelt5e});
        \draw [->] ({pic cs:gdelt6s}) to ({pic cs:gdelt6e});
        \draw [->] ({pic cs:gdelt7s}) to ({pic cs:gdelt7e});
        \draw [->] ({pic cs:gdelt8s}) to ({pic cs:gdelt8e});
    \end{tikzpicture}
    \label{tab:rt}
\end{table*}

\subsection{Experimental Setup}
\noindent \textbf{Datasets.}
We evaluate the performance of \HASTE on five dynamic graph datasets, whose statistics are shown in Table~\ref{tab:ds}.
Among them, \textit{Wikipedia}~\cite{JODIE}, \textit{Reddit}\footnote{The Reddit dataset used in this paper is obtained exclusively from the work~\cite{JODIE}, and no data is directly scraped from the Reddit website.}~\cite{JODIE}, and \textit{MovieLens}~\cite{movielens} are bipartite graphs without node features.
\textit{Flights}~\cite{poursafaei2022towards} is a traffic graph without edge features, and \textit{GDELT}~\cite{tgl} is a large-scale knowledge graph including both node and edge features.
The tasks are to predict user posts (\textit{Wikipedia}, \textit{Reddit}, \textit{MovieLens}), flight  schedules (\textit{Flights}), and news (\textit{GDELT}).
To simulate the use cases in real-world applications, for large-scale datasets with more than one million temporal edges, we use the latest one million edges with 60\%, 20\%, and 20\% chronological splits as the training, validation, and test sets, respectively.

\noindent \textbf{TGNN Models.}
We build \HASTE on two state-of-the-art TGNN models introduced in Section~\ref{sec:bg-tgnn}. 
TGAT~\cite{tgat} uses a 2-layer attention-based temporal aggregator with supporting nodes uniformly sampled from the historical neighbors. GraphMixer~\cite{graphmixer} uses a single-layer MLP-Mixer temporal aggregator with the most recent neighbors as supporting nodes.
To ensure a fair comparison, we keep the number of supporting neighbors to 10, the default value in both baselines. 
Note that \HASTE does not provide any additional input to the TGNN models other than selecting high-quality supporting neighbors of the same size.

\noindent \textbf{Configurations.}
For the TGNN models, we follow the default parameters used in the TGL framework~\cite{tgl} for a fair comparison.
In particular, we use the 0.0001 learning rate, 600 batch size, 200 training epochs, and $n=10$ supporting neighbors per node for all the datasets and all the models.
We set the dimension of all the hidden embeddings and encodings to 100.
For methods with adaptive neighbor sampling, we set $m=25$ as the budget of the neighbor finder for all the datasets, except for the ablation study in Section~\ref{sec:ablation}.
We follow DistTGL~\cite{zhou2023disttgl} to evaluate the performance of transductive temporal link prediction using Mean Reciprocal Rank (MRR) with 49 randomly sampled negative destination nodes.
Please refer to our open-sourced code\footnote{https://github.com/facebookresearch/taser-tgnn} for more details on the hyper-parameters.

\noindent \textbf{Hardware and Software.}
We implement \HASTE using Python 3.11, PyTorch 2.0.1, DGL 1.1, and CUDA 12.2. 
All the experiments are conducted on a machine with dual 96-Core AMD EPYC 9654 CPUs paired with 1.5TB ECC-DDR5 RAM and a single NVIDIA RTX 6000 Ada GPU with 48GB ECC-GDDR6 VRAM.

\subsection{Accuracy} \label{sec:accuracy}
Table~\ref{tab:mrr} shows the accuracy of \HASTE on the five datasets.
We create two variants to better evaluate the effectiveness of each of these two adaptive sampling methods in \HASTE, where \textit{w./ Ada. Mini-Batch} denotes baseline methods with adaptive mini-batch selection and \textit{w./ Ada. Neighbor} is the one with adaptive neighbor sampling.
With both adaptive mini-batch selection and neighbor sampling, \HASTE achieves an average of 2.3\% MRR improvements over the baselines. 
TGAT gets an average of 3.1\% improvements with \HASTE, while GraphMixer only gets 1.4\% improvements.
Intuitively, this is because TGAT takes 2-hop neighbors as the input, which benefits more from the adaptive neighbor sampler compared to the 1-hop neighbors of GraphMixer.
On the one hand, each variant of \HASTE consistently outperforms the baseline TGNNs by a large margin, revealing the effectiveness of \HASTE both in denoising training samples and supporting neighbors.
On the other hand, our results suggest that these two orthogonal adaptive sampling techniques can be employed collectively to either enhance or, at least, maintain accuracy.

We notice that the same neighbor decoder, when paired with different temporal aggregators, leads to remarkably different performances.
This justifies the need for a general encoder-decoder scheme in \HASTE.
In addition, increasing the integrity of the whole model by using a neighbor decoder with a similar architecture as the temporal aggregator can reduce training difficulties and thus improve accuracy.
We note substantial accuracy gains (up to 6\%) when training MLP-Mixer with GraphMixer, yet observe minimal improvements with TGAT, whereas TGAT exhibits a preference for the GATv2 neighbor decoder.
For the neighbor encoder, our proposed frequency encoding and identity encoding consistently work well with any neighbor decoders, reducing the variance of test accuracy and improving the MRR by $0.6\% \sim 1.8\%$.

\subsection{Runtime} \label{sec:runtime}
In this section, we evaluate the speedup of our proposed optimizations in \HASTE.
The training time of \HASTE can be broken down into the four dominant steps: neighbor finding, adaptive neighbor sampling, feature slicing, and forward and backward propagation.
We build \HASTE using the optimized temporal aggregators as proposed in TGL~\cite{tgl}.
For the baseline, we slice all the features from RAM to GPU in each training iteration and use the original neighbor finder implementation in TGAT~\cite{tgat} and GraphMixer~\cite{graphmixer}.
Note that although TGL provides a high-performance parallel CPU neighbor finder, it maintains a pointer array for efficient temporal neighborhood searching that only supports training in chronological order, which does not work in \HASTE since our mini-batch selection is randomly sampled from a dynamic distribution.

\begin{figure*}[t]
    \centering
    \tikzsetnextfilename{linechart1}
\begin{tikzpicture}
    \definecolor{r1}{RGB}{176, 58, 46}
    \definecolor{r2}{RGB}{31, 97, 141}
    \definecolor{r3}{RGB}{17, 122, 101}
    \definecolor{r8}{RGB}{185, 119, 14}

    \begin{groupplot}[
        group style={
            group size=5 by 1,
            horizontal sep=0.3cm,
            vertical sep=0.8cm
        },
        tick label style={font=\scriptsize},
        label style={font=\footnotesize},
        title style={yshift=-1ex, font=\small},
        width=4.57cm,
        height=3.6cm,
        ymode=log, 
        ytick={0.1, 10, 1000}, ymin=0.01, ymax=5000,
        xtick={5,10,15,20,25},
        xlabel={\#neighbors / layer}, xlabel style={yshift=1.5ex},
        legend style={legend columns=3,anchor=center,font=\scriptsize,column sep=0.2cm,xshift=32ex,yshift=-19ex},
    ]
    \node at (-0.5,2.25) {(a)};
    
    \nextgroupplot[title={Wikipedia},ylabel={Time (sec)},ylabel style={yshift=-2.5ex},grid=major,
        ]
    \addplot[color=r1,line width=0.25mm] table[x=scope,y=wiki-orgi,col sep=comma]{data/scope.csv}; \addlegendentry{Origin Neigh Finder (CPU)}
    \addplot[color=r2,line width=0.25mm] table[x=scope,y=wiki-tgl,col sep=comma]{data/scope.csv}; \addlegendentry{TGL Neigh Finder (CPU, only supports chrono. order)}
    \addplot[color=r3,line width=0.25mm] table[x=scope,y=wiki-gpu,col sep=comma]{data/scope.csv}; \addlegendentry{\HASTE Neigh Finder (GPU)}

    \nextgroupplot[title={\textcolor{white}{g}Reddit\textcolor{white}{g}},grid=major,
        yticklabels={},
        ]
    \addplot[color=r1,line width=0.25mm] table[x=scope,y=reddit-orgi,col sep=comma]{data/scope.csv};
    \addplot[color=r2,line width=0.25mm] table[x=scope,y=reddit-tgl,col sep=comma]{data/scope.csv};
    \addplot[color=r3,line width=0.25mm] table[x=scope,y=reddit-gpu,col sep=comma]{data/scope.csv};

    \nextgroupplot[title={\textcolor{white}{g}Flight\textcolor{white}{g}},grid=major,
        yticklabels={},
        ]
    \addplot[color=r1,line width=0.25mm] table[x=scope,y=flight-orgi,col sep=comma]{data/scope.csv};
    \addplot[color=r2,line width=0.25mm] table[x=scope,y=flight-tgl,col sep=comma]{data/scope.csv};
    \addplot[color=r3,line width=0.25mm] table[x=scope,y=flight-gpu,col sep=comma]{data/scope.csv};

    \nextgroupplot[title={\textcolor{white}{g}MovieLens\textcolor{white}{g}},grid=major, 
        yticklabels={},
        ]
    \addplot[color=r1,line width=0.25mm] table[x=scope,y=movielens-orgi,col sep=comma]{data/scope.csv};
    \addplot[color=r2,line width=0.25mm] table[x=scope,y=movielens-tgl,col sep=comma]{data/scope.csv};
    \addplot[color=r3,line width=0.25mm] table[x=scope,y=movielens-gpu,col sep=comma]{data/scope.csv};

    \nextgroupplot[title={\textcolor{white}{g}GDELT\textcolor{white}{g}},grid=major,
        yticklabels={},
        ]
    \addplot[color=r1,line width=0.25mm] table[x=scope,y=gdelt-orgi,col sep=comma]{data/scope.csv};
    \addplot[color=r2,line width=0.25mm] table[x=scope,y=gdelt-tgl,col sep=comma]{data/scope.csv};
    \addplot[color=r3,line width=0.25mm] table[x=scope,y=gdelt-gpu,col sep=comma]{data/scope.csv};
        
    \end{groupplot}

    \draw[<->, densely dashed, line width=0.3mm, orange] (2.74,0.78) -- (2.74,0.15)
        node[midway, left] {\small 43$\boldsymbol{\times}$};
    \draw[<->, densely dashed, line width=0.3mm, orange] (6.02,1.05) -- (6.03,0.41)
        node[midway, left] {\small 52$\boldsymbol{\times}$};
    \draw[<->, densely dashed, line width=0.3mm, orange] (9.31,1.08) -- (9.31,0.43)
        node[midway, left] {\small 56$\boldsymbol{\times}$};
    \draw[<->, densely dashed, line width=0.3mm, orange] (12.60,1.14) -- (12.60,0.51)
        node[midway, left] {\small 42$\boldsymbol{\times}$};
    \draw[<->, densely dashed, line width=0.3mm, orange] (15.89,1.17) -- (15.89,0.57)
        node[midway, left] {\small 37$\boldsymbol{\times}$};
\end{tikzpicture}

\vspace*{0.1cm}

\tikzsetnextfilename{linechart2}
\begin{tikzpicture}
    \definecolor{r1}{RGB}{176, 58, 46}
    \definecolor{r2}{RGB}{31, 97, 141}
    \definecolor{r3}{RGB}{17, 122, 101}
    \definecolor{r8}{RGB}{0, 0, 0}

    \begin{groupplot}[
        group style={
            group size=4 by 1,
            horizontal sep=0.68cm,
            vertical sep=0cm
        },
        tick label style={font=\scriptsize},
        label style={font=\footnotesize},
        width=5.1cm,
        height=3.6cm,
        xlabel={Epoch},xlabel style={yshift=1.5ex},
        title style={yshift=-1ex, font=\small},
        legend style={legend columns=4,anchor=center,font=\scriptsize,column sep=0.2cm,xshift=29ex,yshift=-19ex},
    ]
    \node at (-0.5,2.4) {(b)};
    \nextgroupplot[title={Wikipedia},ylabel={Cache Hit Rate (\%)}, ylabel style={yshift=-2ex}, grid=major,
    ymin=65,ymax=95,
    ]
    \addplot[color=r8,line width=0.25mm] table[x=epoch,y=owiki_0.1,col sep=comma]{data/cache.csv};     
    \addplot[color=r1,line width=0.25mm] table[x=epoch,y=wiki_0.1,col sep=comma]{data/cache.csv}; 
    \addplot[color=r2,line width=0.25mm] table[x=epoch,y=wiki_0.2,col sep=comma]{data/cache.csv};
    \addplot[color=r3,line width=0.25mm] table[x=epoch,y=wiki_0.3,col sep=comma]{data/cache.csv};
    \addlegendentry{Oracle Cache  $10\%, 20\%, 30\%$}
    \addlegendentry{\HASTE Cache 10\%}
    \addlegendentry{\HASTE Cache 20\%}
    \addlegendentry{\HASTE Cache 30\%}
    \addplot[color=r8,line width=0.25mm] table[x=epoch,y=owiki_0.2,col sep=comma]{data/cache.csv}; 
    \addplot[color=r8,line width=0.25mm] table[x=epoch,y=owiki_0.3,col sep=comma]{data/cache.csv};
  
    \nextgroupplot[title={Reddit},grid=major,
        ymin=65,ymax=95,
        ]
       \addplot[color=r1,line width=0.25mm] table[x=epoch,y=reddit_0.1,col sep=comma]{data/cache.csv};
    \addplot[color=r8,line width=0.25mm] table[x=epoch,y=oreddit_0.1,col sep=comma]{data/cache.csv};
        \addplot[color=r2,line width=0.25mm] table[x=epoch,y=reddit_0.2,col sep=comma]{data/cache.csv};
    \addplot[color=r8,line width=0.25mm] table[x=epoch,y=oreddit_0.2,col sep=comma]{data/cache.csv};
        \addplot[color=r3,line width=0.25mm] table[x=epoch,y=reddit_0.3,col sep=comma]{data/cache.csv};
    \addplot[color=r8,line width=0.25mm] table[x=epoch,y=oreddit_0.3,col sep=comma]{data/cache.csv};

    \nextgroupplot[title={MovieLens},grid=major,
        ymin=55,ymax=85,
        ]
    \addplot[color=r1,line width=0.25mm] table[x=epoch,y=movielens_0.1,col sep=comma]{data/cache.csv};
    \addplot[color=r8,line width=0.25mm] table[x=epoch,y=omovielens_0.1,col sep=comma]{data/cache.csv};
    \addplot[color=r2,line width=0.25mm] table[x=epoch,y=movielens_0.2,col sep=comma]{data/cache.csv};
    \addplot[color=r8,line width=0.25mm] table[x=epoch,y=omovielens_0.2,col sep=comma]{data/cache.csv};
    \addplot[color=r3,line width=0.25mm] table[x=epoch,y=movielens_0.3,col sep=comma]{data/cache.csv};
    \addplot[color=r8,line width=0.25mm] table[x=epoch,y=omovielens_0.3,col sep=comma]{data/cache.csv};

    \nextgroupplot[title={GDELT},grid=major,
        ymin=85,ymax=101
        ]
    \addplot[color=r1,line width=0.25mm] table[x=epoch,y=gdelt_0.1,col sep=comma]{data/cache.csv};
    \addplot[color=r8,line width=0.25mm] table[x=epoch,y=ogdelt_0.1,col sep=comma]{data/cache.csv};
    \addplot[color=r2,line width=0.25mm] table[x=epoch,y=gdelt_0.2,col sep=comma]{data/cache.csv};
    \addplot[color=r8,line width=0.25mm] table[x=epoch,y=ogdelt_0.2,col sep=comma]{data/cache.csv};
    \addplot[color=r3,line width=0.25mm] table[x=epoch,y=gdelt_0.3,col sep=comma]{data/cache.csv};
    \addplot[color=r8,line width=0.25mm] table[x=epoch,y=ogdelt_0.3,col sep=comma]{data/cache.csv};
    \end{groupplot}
\end{tikzpicture}%
    \caption{(a) Total sampling time per epoch of a $2$-layer TGAT with different neighbor finders and different numbers of neighbors per layer. (b) Cache Hit Rate of \HASTE caching strategy and Oracle caching strategy with different training epochs.}
    \label{fig:scope+cache}
\end{figure*}
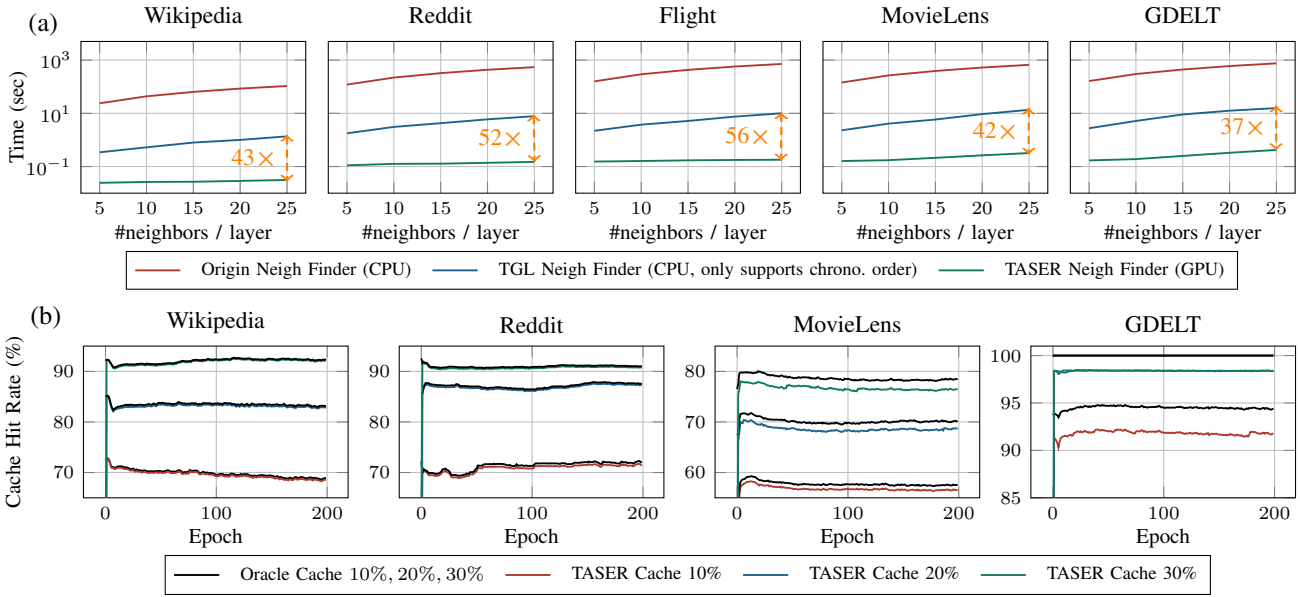

As shown in Table~\ref{tab:rt}, the bottlenecks of the baseline are neighbor finding and feature slicing. 
\modify{
After applying our GPU neighbor finder and GPU feature caching with 20\% of total edge features, the ratio of mini-batch generation time (i.e., neighbor finding time plus feature slicing time) to the total runtime drops significantly from $40\% \sim 92\%$ to $3\% \sim 18\%$.
}
The rest of the runtime mainly lies in the neural network computation, which is proportional to the computational complexity.
Since the \textit{Flights} dataset does not contain edge features and the node features can be entirely stored on GPU, we do not demonstrate its runtime.
\HASTE achieves an average of 8.68$\times$ speedup on TGAT and 1.77$\times$ speedup on GraphMixer.
TGAT is a 2-layer TGNN and requires a squared number of supporting neighbors, suffering a greater impact from the inefficiency of neighbor finding and feature slicing.
On \textit{GDELT}, since we use the latest one million temporal edges for training and evaluation, caching 20\% of edge features is already sufficient for the training set.

\subsection{GPU Neighbor Finder}
Fig.~\ref{fig:scope+cache}(a) compares the runtime of different uniform neighbor finders, including the original Python-implemented neighbor finder~\cite{tgat}, the high-performance CPU parallel neighbor finder from TGL~\cite{tgl}, and our \HASTE GPU neighbor finder. 
Since the TGL neighbor finder only supports chronological training order, we use chronological order on all three neighbor finders for a fair comparison. 
To better reflect the actual runtime of CPU neighbor finders, we also include the CPU-GPU data loading time for the sampled neighbor indices.
Note that the TGL neighbor finder is built on a pointer array that leverages the chronological training order for fast memory access. Although we do not specifically optimize for the chronological training order, our GPU neighbor sampler is still orders of magnitude faster than the TGL neighbor finder.
As shown in Fig.~\ref{fig:scope+cache}(a), when the number of neighbors per layer is set to 25, our \HASTE neighbor finder achieves a speedup of more than three orders of magnitude compared to the original neighbor finder, and a $37 \sim 56\times$ speedup compared to the TGL neighbor finder, across all five datasets. 

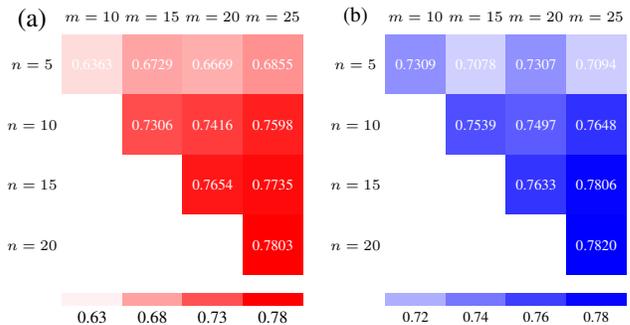
\begin{figure}[b!]
    \centering
    \tikzsetnextfilename{heatmap}
\begin{tikzpicture}
    \node at (0,-0.2) {(a)};
    \begin{scope}[scale=0.8, every node/.append style={transform shape}]
        \foreach \y [count=\n] in 
        {
            {0.6363, 0.6729, 0.6669, 0.6855},
            {, 0.7306, 0.7416, 0.7598},
            {, , 0.7654, 0.7735},
            {, , ,0.7803},
        } 
        {
            \foreach \x [count=\m] in \y 
            {
                \pgfmathparse{100+(\x-0.7803)*600}
                \let\myopacity\pgfmathresult
                \node[fill=red!\myopacity,minimum size=1cm,text=white] at (\m,-\n) {\scriptsize \x};
            }
        }
                
        \foreach \a [count=\i] in {10,15,20,25} 
        {
            \node[minimum size=6mm] at (\i,-0.2) {\scriptsize $m=\a$};
        }
    
        \foreach \a [count=\i] in {5,10,15,20} 
        {
            \node[minimum size=6mm] at (0,-\i) {\scriptsize $n=\a$};
        }

        \foreach \x [count=\i] in {0.63,0.68,0.73,0.78}
        {
            \pgfmathparse{100+(\x-0.7803)*630}
            \let\myopacity\pgfmathresult
            \path[fill=red!\myopacity] (\i-0.5,-5) rectangle ++(1,0.2);
            \node[] at (\i,-5.2) {\footnotesize \x};
        }
    \end{scope}

    \begin{scope}[shift={(4.3,0)},scale=0.8, every node/.append style={transform shape}]
        \node at (0,-0.2) {(b)};
        \foreach \y [count=\n] in 
        {
            {0.7309, 0.7078, 0.7307, 0.7094},
            {, 0.7539, 0.7497, 0.7648},
            {, , 0.7633, 0.7806},
            {, , , 0.7820},
        } 
        {
            \foreach \x [count=\m] in \y 
            {
                \pgfmathparse{100+(\x-0.782)*1100}
                \let\myopacity\pgfmathresult
                \node[fill=blue!\myopacity,minimum size=1cm,text=white] at (\m,-\n) {\scriptsize \x};
            }
        }
                
        \foreach \a [count=\i] in {10,15,20,25} 
        {
            \node[minimum size=6mm] at (\i,-0.2) {\scriptsize $m=\a$};
        }

        \foreach \a [count=\i] in {5,10,15,20} 
        {
            \node[minimum size=6mm] at (0,-\i) {\scriptsize $n=\a$};
        }

        \foreach \x [count=\i] in {0.72,0.74,0.76,0.78}
        {
            \pgfmathparse{100+(\x-0.782)*1100}
            \let\myopacity\pgfmathresult
            \path[fill=blue!\myopacity] (\i-0.5,-5) rectangle ++(1,0.2);
            \node[] at (\i,-5.2) {\scriptsize \x};
        }
    \end{scope}

\end{tikzpicture}
    \caption{Test MRR of (a) TGAT and (b) GraphMixer with \HASTE on the Wikipedia dataset. $m$ and $n$ denote the numbers of neighbors selected by the neighbor finder and the adaptive neighbor sampler, respectively.}
    \label{fig:ablation}
\end{figure}

\subsection{GPU Cache}
We compare our GPU caching strategy with the Oracle caching strategy, which assumes the access frequency of each edge is known in advance. Both our caching strategy and the Oracle caching strategy are updated at the end of each epoch.
Fig.~\ref{fig:scope+cache} (b) shows that our GPU caching strategy achieves a near-optimal cache hit rate, close to the Oracle cache with the same size.
We choose the 10\%, 20\%, and 30\% cache ratio as they fit mainstream GPUs with 8GB, 16GB, and 40GB VRAM on the GDELT dataset, respectively.
The cache hit rates increase proportionally with the cache ratio until the Oracle cache is able to include all the accessed features.
We observe that, as the entire model's weights progressively stabilize, our GPU cache rarely necessitates an update after 20 epochs, further reaffirming the low maintenance of our strategy.
Note that the cache hit rate of the Oracle caching strategy can also reflect the explore-and-exploit strategy of the adaptive samplers. 
For instance, on the \textit{Wikipedia} dataset, the cache hit rate of the 10\% Oracle cache first increased to 73\% and then gradually decreased to 69\%, illustrating that the adaptive samplers initially exploit high-reward edges and subsequently explore other training samples and supporting neighbors to improve the accuracy.

\subsection{Ablation Study} \label{sec:ablation}
We evaluate the performance of \HASTE with different neighbor budgets. Fig.~\ref{fig:ablation} demonstrates that \HASTE is versatile to various numbers of neighbor candidates $m$ and sampled supporting neighbors $n$. 
We note that the accuracy does not improve when increasing $m$ for GraphMixer with $n=5$.
Since GraphMixer is a one-layer TGNN model, it has only $5$ supporting nodes per root node when $n=5$, while a 2-layer TGAT has 
$5+5\times5=30$ supporting nodes. 
Selecting $n=5$ as the hyper-parameter choice for GraphMixer is suboptimal for real-world applications, leading to inaccurate supervision for the adaptive sampler from the TGNN model.
The results validate our hypothesis that, with a larger number of neighbor candidates $m$, the adaptive neighbor sampler is capable of selecting supporting neighbors that provide more pivotal information for task prediction.
It also shows that \HASTE consistently performs well when TGNNs prefer a larger number of supporting neighbors $n$.

\section{Conclusion}
In this work, we proposed \HASTE, a novel temporal neighbor sampling method for fast and accurate representation learning on dynamic graphs.
With a two-fold adaptive sampling method, temporal adaptive mini-batch selection, and temporal adaptive neighbor sampling, \HASTE endowed TGNNs with the ability to handle distinctive noise in dynamic graphs.
\modify{
However, the introduced adaptive sampling increases the number of neighborhood traversals, prolonging the runtime of existing bottlenecks in TGNNs, specifically neighbor finding and CPU-GPU feature slicing.
We proposed two system optimizations to address these bottlenecks: an efficient GPU neighbor finder and a GPU feature caching strategy.
}
On two state-of-the-art backbone TGNNs and five real-world datasets, \HASTE not only improved the accuracy by an average of 2.3\% in MRR but also achieved an average speedup of 5.1$\times$ on a single GPU.

\section*{Acknowledgment}
This work is supported by Meta Platforms Inc. under grant number INB2675366, National Science Foundation (NSF) under grant OAC-2209563, and DEVCOM Army Research Lab (ARL) under grant W911NF2220159. 
Distribution Statement A: Approved for public release.
Distribution is unlimited.

\bibliographystyle{IEEEtran}
\bibliography{cite}

\end{document}